\documentclass[conf]{new-aiaa}
\usepackage[utf8]{inputenc}

\usepackage[version=4]{mhchem}
\usepackage{longtable,tabularx}
\usepackage{amsmath, bbm}
\usepackage{booktabs}
\usepackage{multirow}
\usepackage{cleveref}
\usepackage{graphicx}
\usepackage[ruled,vlined]{algorithm2e}
\usepackage{siunitx}

\usepackage{booktabs,caption}
\usepackage[flushleft]{threeparttable}

\usepackage[export]{adjustbox}

\title{Obstacle Avoidance Using a Monocular Camera}

\author{Kyle Hatch\footnote{Undergraduate Student, Department of Computer Science, khatch@stanford.edu}, John Mern\footnote{Ph.D. Candidate, Department of Aeronautics and Astronautics, AIAA Student Member}, and Mykel Kochenderfer\footnote{Associate Professor, Department of Aeronautics and Astronautics, AIAA Associate Fellow}} 
\affil{Stanford University, Stanford, CA, 94305}

\begin{document}

\maketitle

\begin{abstract}

A collision avoidance system based on simple digital cameras would help enable the safe integration of small UAVs into crowded, low-altitude environments. 
In this work, we present an obstacle avoidance system for small UAVs that uses a monocular camera with a hybrid neural network and path planner controller. 
The system is comprised of a vision network for estimating depth from camera images, a high-level control network, a collision prediction network, and a contingency policy. 
This system is evaluated on a simulated UAV navigating an obstacle course in a constrained flight pattern. 
Results show the proposed system achieves low collision rates while maintaining operationally relevant flight speeds.
\end{abstract}




\section{Introduction}
The safe introduction of unmanned vehicles into congested urban environments will require robust obstacle sensing and avoidance systems. 
Small unmanned aerial vehicles (UAVs) often have size, weight, and power consumption (SWaP) constraints on the payload, making traditional obstacle avoidance sensors such as radar infeasible~\cite{Sahawneh2016}.
Cameras are light-weight and low-power and are commonly found on UAVs.
However, cameras do not directly detect the distance of observed objects, making their use in conventional obstacle avoidance systems challenging.
Methods to estimate depth from images and provide robust obstacle avoidance commands are needed for low-SWaP systems.

Vision-based obstacle avoidance is challenging due to the difficulties of estimating obstacle depth from 2D images~\cite{ng2005car, mori2013firstresults}.
Deep neural network vision systems have been able to improve upon conventional depth estimation methods, however the resulting estimates remain noisy~\cite{mern2019}.
Deep reinforcement learning has been shown to be effective for high-level control with noisy, high-dimensional inputs~\cite{levine2016}.
Training a network to 
pilot a vehicle to a destination without collisions has challenged standard learning approaches~\cite{pirotta2015multi}.


The purpose of this work is to develop a camera-based neural network guidance and collision avoidance system.
Obstacle detection systems estimate the distance of objects in a vehicle's field of view (FoV) and predict the likelihood of a collision based on those estimates and the currently planned vehicle trajectory. 
Monocular camera images do not explicitly contain range information, making their direct use in obstacle detection challenging.

Several methods exist to estimate depth from single images or image sequences~\cite{ng2005car,lee2011smalluavs}. 
Conventional techniques such as depth-from-motion or context-based estimation are highly sensitive to scene content and may produce large errors~\cite{natraj2014realtime}.
Convolutional Neural Networks (CNNs) may be more robust to scene type, but typically result in noisy estimates~\cite{mern2019}, making them difficult to use with conventional path planning methods.


Deep Reinforcement Learning (RL) can effectively solve high-dimensional control problems but it can be ineffective for multi-objective or constrained problems~\cite{dewey2014reinforcement}. 
This is in part because neural network behavior is highly sensitive to the choice of reward function, which serves as the optimization target during training. 
The task of piloting a UAV to a goal destination while avoiding obstacles is inherently a multi-objective problem: the task of reaching the goal destination in as little time as possible must be balanced with the task of minimizing the risk of collision. 
Tuning the reward such that the network learns to solve the target task while avoiding failures is difficult and incorrect parameters can lead to undesirable behavior.
 
Path planning algorithms such as $A^*$~\cite{hart68} can safely navigate UAVs around obstacles while minimizing the risk of collision given an accurate map. 
With map uncertainty, path planners often rely on reachability analysis with large error margins to ensure safe operation~\cite{Herbert19}. 
This can lead to overly conservative behavior where generated vehicle paths are much longer than optimal.
One approach to overcome these problems is to develop a hybrid system in which specialized sub-systems solve the path planning and obstacle avoidance sub-tasks and enforce behavior constraints~\cite{ding2011}. 

We propose a hybrid neural network, path planner control system that can solve the multi-objective collision avoidance problem. 
The system is composed of three neural networks and a contingency policy.
Each of the neural networks performs a distinct task: one network is used for depth estimation, one for collision prediction, and one for vehicle control.
The contingency policy is either a path planning based algorithm such as $A^*$ or an expert policy.
During unobstructed flight, the vehicle control neural network pilots the aircraft quickly towards a goal destination. 
At each timestep, the collision detection network estimates the probability that the vehicle will collide with an obstacle within a receding horizon window. 
If the collision probability exceeds a threshold, control of the aircraft is temporarily given to the contingency policy. 
The contingency policy safely pilots the aircraft around the obstacle and then returns control of the aircraft to the learned neural network controller. The trained system is evaluated on its ability to efficiently reach a goal destination while avoiding stationary obstacles.

By decomposing the multi-objective collision avoidance problem into these two modes of operation, our hybrid system leverages the advantages offered by both the neural network controller and the contingency policy.
The neural network controller is trained using RL and can learn a wide range of behaviors that enable it to efficiently pilot the aircraft towards a destination. 
Using the vehicle control network as the default controller of the aircraft avoids using a highly conservative path planning algorithm when such caution is unnecessary. 
This enables the hybrid system pilot the aircraft to its destination more quickly than if the contingency policy were to control the aircraft the entire time.
However, in high risk situations a conservative contingency policy can pilot the aircraft around nearby obstacles more safely than the learned RL policy can. 
Because the contingency policy is used exclusively for avoiding obstacles that are very close, it can be optimized to perform this single, specific purpose. 
In this way, our system balances the dual objectives of quickly flying to a goal destination and avoiding collisions without requiring extensive tuning of the RL reward function. 

\section{Prior Work}
Camera-based collision avoidance systems have been developed using a variety of methods~\cite{kochenderfer2008}. 
Some systems rely on visual cues to estimate depth from monocular images. 
\Citeauthor{ng2005car}~\cite{ng2005car} analyze texture queues from monocular images to estimate the depth of the closest object in a series of discrete regions within the image. 
They then train a model-based RL agent to drive a small remote control car around stationary obstacles at speeds of up to 5 meters per second. 
\Citeauthor{mori2013firstresults}~\cite{mori2013firstresults} estimate depth by analyzing the change in relative size of image patches between frames. 
Identifying object features, such as corners and edges, and comparing the size and location of these features between frames is also a common approach for monocular depth estimation~\cite{lee2011smalluavs, natraj2014realtime, alkaff2016detection/avoidance, kovacs2016visual}. 
While approaches using visual cues can be successful at estimating depth in environments where specific visual cues are present, they tend to generalize poorly to other types of environments. 

Stereoscopic techniques have also been used to estimate depth in images for obstacle avoidance systems. 
In these approaches, two cameras located at a known distance apart from each other are used, and the disparity between the two images is used to estimate depth \cite{Lucas:1981:IIR:1623264.1623280}. 
Such approaches have been demonstrated to work with collision avoidance systems for autonomous aircraft \cite{Ruf_2018}. 
However, the distance at which they are effective is limited by how far apart the two cameras are positioned from each other~\cite{Alvertos1988ResolutionLA}, limiting their utility on small platforms.

More recently, CNNs have been shown to be effective at estimating depth from monocular images in a variety of environments~\cite{laina2016deeper}.
Several different CNN architectures have been developed specifically for the purpose of being used by obstacle avoidance systems for autonomous aircraft~\cite{Mancini_2018, chakravarty2017cnnbased, mern2019}. 

In addition to obtaining accurate depth mappings of their surroundings, obstacle avoidance systems must be able to use these maps to avoid collisions with obstacles.
Path planning algorithms are often used with CNN depth estimators to maneuver aircraft around detected obstacles. 
Although path planning algorithms can be effective, they typically require tuning a large number of parameters and do not learn new behaviors from experience~\cite{xie2017monocular}. 

Reinforcement learning methods that learn to avoid obstacles through trial and error without supervision or expert knowledge have also been proposed.
An agent learns from experience to find mappings from states to actions that maximize a long term, accumulated reward. 
RL has been shown to be successful in a wide variety of areas, including video games~\cite{mnih2013playing} and robotic control~\cite{amarjyoti2017deep}. 
Similar simulation-based methods have been successfully applied to specific obstacle avoidance tasks as well~\cite{kochenderfer2008, temizer2010}.

Several previous works have addressed the issue of developing camera based obstacle avoidance systems using RL.
\Citeauthor{xie2017monocular}~\cite{xie2017monocular} use a convolutional residual neural network to produce pixel-wise depth estimates from RGB images. 
They then train a deep double Q-network to pilot a small wheeled robot around obstacles. \Citeauthor{anwar2018navrenrl}~\cite{anwar2018navrenrl} also use double deep Q-networks to train an agent to pilot a small quadrotor around stationary obstacles. 
For depth estimation, they use the network architecture developed by \Citeauthor{laina2016deeper}~\cite{laina2016deeper}. 
Similarly, \Citeauthor{singla2018memorybased} ~\cite{singla2018memorybased} use deep Q-learning with attention to pilot a quadrotor around obstacles in a simulator using discrete directional commands. 

Recently, hybrid systems that utilize a combination of learning and planning methods have been proposed for general vehicle guidance~\cite{dhinakaran17}.
These systems are designed to use the efficient route planning of a learned system during nominal operation and use more conservative contingency planning during high-risk situations. 
At the time of this work, no hybrid guidance system has yet been proposed using only monocular images for obstacle detection.

These previously approaches have some important limitations. 
Previously proposed systems have been trained and tested in indoor environments. 
Indoor environments are visually much simpler than outdoor environments, which makes depth estimation easier. 
The use of an action space consisting of moving forward a fixed distance or rotating right or left by a fixed amount, as used by \Citeauthor{singla2018memorybased}~\cite{singla2018memorybased} and \Citeauthor{anwar2018navrenrl}~\cite{anwar2018navrenrl}, limit the potential maneuvers the aircraft can perform.
These obstacle avoidance systems must operate at relatively slow speeds because of their reliance on stop-and-go flight maneuvers. 
These limitations decrease the usefulness of these systems on aircraft flying practical missions. 
In contrast, our system operates in a visually complex outdoor environment at comparatively high speeds without the use of stop-and-go flight maneuvers. 

\section{Methods}
\begin{figure}[ht]
    \centering
    \includegraphics[width=0.99\textwidth]{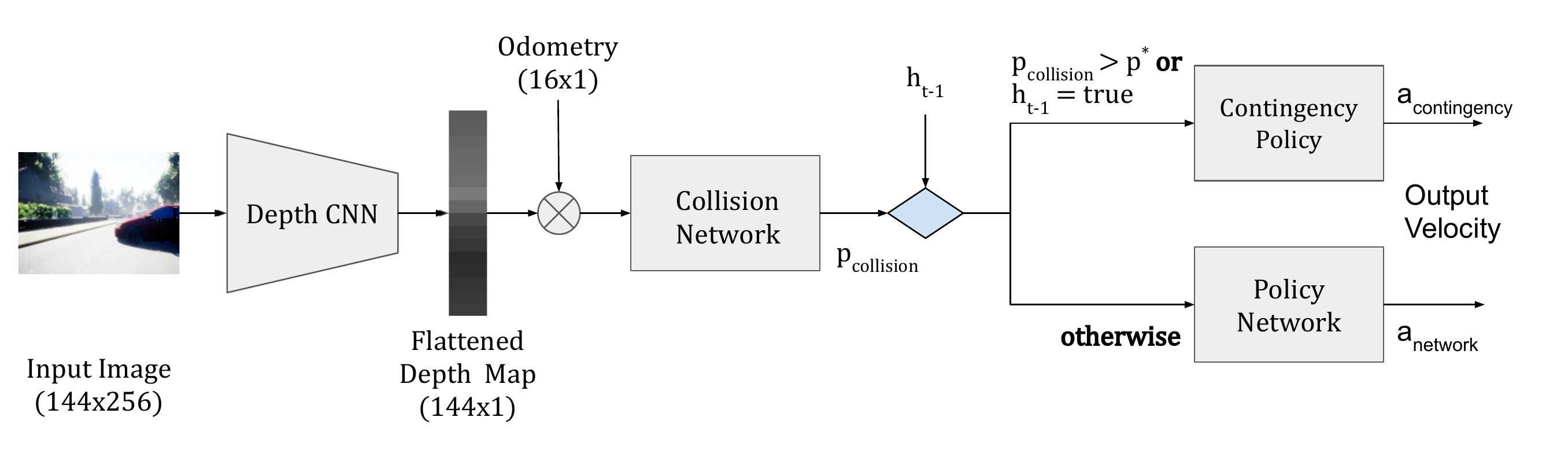}
    \caption{System diagram: This figure shows the information flow through the collision avoidance system. The RGB input images are processed by the depth CNN and the reduced-scale output is flattened into a 144-length vector. 
    The depth vector is concatenated with the vehicle odometry vector and input into the collision prediction network, which outputs the probability of collision.
    If the collision probability is below a target threshold and the contingency policy is not currently executing, the input vector is passed to the policy network, which outputs a velocity command.
    Otherwise, the contingency policy is queried. 
    Once activated, the contingency policy pilots the aircraft for a short period of time until it terminates. 
    Control of the aircraft is then returned to the policy network. 
    }
    \label{fig:system}
\end{figure}
The inputs to the proposed collision avoidance system are images from an on-board visual-spectrum camera and filtered kinematic estimates from the vehicle inertial sensors. 
The system outputs a velocity command, which is used as input to a low-level controller. 
The collision avoidance system is comprised of three neural networks--a depth map estimator CNN, a control policy Q-network, and a collision prediction network--and a non-learned contingency policy.
An architecture diagram is shown in~\cref{fig:system}.

\begin{figure}
    \centering
    \includegraphics[width=0.3\textwidth]{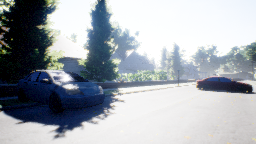}
    \includegraphics[width=0.3\textwidth]{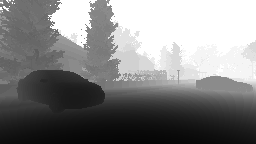}
    \includegraphics[width=0.3\textwidth]{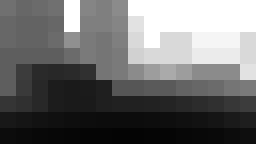}
    \caption{Depth map example: (Left) Input RGB image. (Center) Full resolution depth map. (Right) 1/16 resolution depth map. Pixel intensity is proportional to the estimated distance. Darker pixels indicate closer objects.}
    \label{fig:depth_maps}
\end{figure}

The depth CNN estimates a depth map from the on-board camera images. 
A depth map is an array of distance estimates where each value in the array corresponds to a location in the original field-of-view. 
In a full resolution depth map (where the size of the depth map is equal to the size of the input image), the value of each depth map pixel is the 
estimated distance from the camera to the object seen at the input image pixel. 
The depth CNN outputs a 1/16 resolution depth map, where each value is an estimate of the minimum distance of the objects in the corresponding $16\times16$ pixel block in the original image. 
An example of depth maps at full and reduced resolution is shown in~\cref{fig:depth_maps}.
The system is run at 10 Hz. 

The depth map estimates are flattened into vectors and concatenated with a kinematics estimate provided by the on-board odometry sensors. 
This joint vector is input to the policy network. 
Based on this input, the policy network chooses an action from a discrete set of predefined velocity commands. 
It does so by estimating the expected value of taking each action in this set and recommending the action with the highest expectation.


The joint input vector is also provided to the collision prediction network as input. 
The collision prediction network estimates the probability that a collision will occur in the next 10 time-steps (corresponding to 1 second at the frequency used in this work). 
If the predicted collision probability exceeds some specified threshold (nominally $50\%$), a contingency policy is enacted. 
This contingency policy temporarily takes control of the aircraft from the RL policy in order to pilot the aircraft around the sensed obstacle(s). 
Once the contingency policy terminates, control of the aircraft is returned to the RL policy. 


In this work, we tested both a rules-based policy and an online path planner as the contingency policy. 
The first policy is an online path planner based on the $A^*$ algorithm. 
As an input to $A^*$, a 2D obstacle map was generated using the last depth estimate frame.
$A^*$ was then executed until a route around the obstacle was found. The details of how the obstacle maps were generated and how $A^*$ was used are described in the Appendix.
The second policy is a simple expert policy designed to pilot the aircraft around an obstacle in front of it. 
The expert policy is also described in the Appendix.
These contingency policies worked well for the experiments we conducted,  however, the policies can be replaced by any suitable policy or planner. 


A prototype system was developed to control a small UAV in Microsoft's AirSim~\cite{shah2017airsim} simulator. 
The host vehicle was a simple quadrotor with a forward mounted monocular camera. 
The training images and true depth maps were gathered as matched-pairs from AirSim.
Velocity commands were input to the AirSim velocity controller API. 
The testing was conducted in the ``Neighborhood'' simulated environment. 

\subsection{Depth Network}
The depth network is a down-scaling convolutional neural network.
Inputs to the network are images represented as 3D arrays of dimension $H\times W \times 3$, where 3 is the number of color channels in the RGB representation.
The network architecture is detailed in~\cref{tab:depth_net}. 
Each layer consists of a standard affine 2D convolution by the layer filters, followed by a non-linear operation. 
In this case, we used the leaky rectified linear unit (Leaky-ReLU) as the non linear activation function.
The leaky-ReLU tends to suffer fewer issues with vanishing gradients than the basic ReLU operation~\cite{leakyrelu}. 

We explored using sequences of images as input to the network to enable depth-from-motion estimation within the graph. 
In this setting, to estimate a depth map $y_t$, the network would have the sequence of $b$ prior images $(x_{t-b}, \ldots, x_t)$, as an input, concatenated by the channel axis such that the complete input would have dimensions $H \times W\times 3b$.
Our initial results show that there are no significant performance benefits to the complete system using this approach. 
The remaining work was conducted using single image inputs to reduce computation. 
\begin{table}[ht]
    \centering
    \caption{Depth Network Architecture. Input dimensions are shown for input sequences of $b$ images, where $b=1$ corresponds to single image inputs.}
    \begin{tabular}{@{}ccccccc@{}}
         Layer & Input Dims & Transform & Kernel Size & Stride & Activation & Output Dims\\
         \toprule
         Input & $H\times W\times 3b $ & Conv2D &  5 & 2 & Leaky ReLU & $H/2\times W/2\times 32 $ \\
         \midrule
         H1 & $H/2\times W/2\times 32 $ & Conv2D &  5 & 2 & Leaky ReLU & $H/4\times W/4\times 64 $ \\
         H2 & $H/4\times W/4\times 64 $ & Conv2D &  3 & 2 & Leaky ReLU & $H/8\times W/8\times 128 $ \\
         H3 & $H/8\times W/8\times 128 $ & Conv2D &  3 & 2 & Leaky ReLU & $H/16\times W/16\times 256 $ \\
         H4 & $H/16\times W/16\times 256 $ & Conv2D &  3 & 1 & Leaky ReLU & $H/16\times W/16\times 128 $ \\
         H5 & $H/16\times W/16\times 128 $ & Conv2D &  3 & 1 & Leaky ReLU & $H/16\times W/16\times 32 $ \\
         \midrule
         Output & $H/16\times W/16\times 32 $ & Conv2D &  3 & 1 & None & $H/2\times W/2\times 1 $ \\
         \bottomrule
    \end{tabular}
    \label{tab:depth_net}
\end{table}

The network was initially trained using a data set of matched image and depth map pairs. 
The training data set was gathered by manually piloting the aircraft through the AirSim environment and recording the camera view and depth map at a 10 Hz.
The network was trained to minimize the expected Huber loss between the true depth map $y_i$ and the depth map estimate produced by the network $\hat{y}_i$. 
To produce the depth map targets, the full-resolution depth  maps were down-scaled using a min-pooling operation, where the value of each element in the $H/16 \times W/16$ array was set to the minimum value of the corresponding $16 \times 16$ block in the full resolution map. 
The expected loss $L(x_i, y_i)$ was estimated by the mean loss over a batch of $m$ training images as
\begin{equation}
    \hat{L}(x_i, y_i) = \frac{1}{m}\sum_i^m\textsc{Huber}(\phi(x_i), y_i)
\end{equation}
where $\phi(x_i) = \hat{y}_i$ denotes the depth network output from the input image $x_i$. 
The Huber loss for a given image pair is defined as
\begin{equation}
    \textsc{Huber}(\hat{y}_i, y_i) = \frac{1}{n}\sum_j^n
    \begin{cases}
    \frac{1}{2} (y_i^j - \hat{y}_i^j), & \text{for} |y_i^j - \hat{y}_i^j| \leq \delta \\
    \delta|y_i^j - \hat{y}_i^j| - \frac{1}{2}\delta^2,               & \text{otherwise}
\end{cases}
\end{equation}
where $j$ is the array element index and $\delta$ is some constant hyper-parameter. 
In this work, $\delta$ was set to $1$.
The input and output images were normalized prior to training such that all values were within $[-1.0, 1.0]$.

\subsection{Policy Network}
The policy network was implemented as a Deep Q-Network (DQN) using the extensions recommended in the RAINBOW architecture~\cite{hessel2018rainbow}.
The network takes as input a vector of the flattened depth map estimate concatenated with a kinematic estimate from the on-board vehicle sensors. 
The kinematic state estimate is the vector $(p_y, v_x, v_y, v_z, a_x, a_y, a_z, \phi, \gamma, \theta, \dot{\phi}, \dot{\gamma}, \dot{\theta}, \ddot{\phi}, \ddot{\gamma}, \ddot{\theta})$, which correspond to the lateral position, linear velocity, linear accelerations, angular orientations, angular rates, and angular accelerations respectively. 
The lateral position $p_y$ was included in this vector so that the policy network would be able to learn not to pilot the quadrotor out of bounds. 
The inclusion of $p_y$ cannot aid in obstacle avoidance, as the policy network was trained on a course in which obstacles are randomly positioned at the beginning of each episode.
The network outputs the Q-values of each action in the action-set. 
The Q-value is defined as the expected sum of discounted rewards when taking a given action and following a fixed policy for the subsequent steps.

The network architecture is a dual-stream network. 
Inputs are pre-processed through three hidden-layers with 128 hidden units each and ReLU activation functions. 
This output is then passed through two parallel streams, each with two hidden layers of 128 units. 
One stream estimates the state value and the other estimates the state-action advantage.
The outputs from each stream are combined and the Q-value for each action is then estimated using a discrete atomized estimator process.
To facilitate exploration during training, a noise term is added to the output of the final layer.
For more details on this architecture and training process, please see the RAINBOW paper~\cite{hessel2018rainbow}.

The action set was defined as the set of four actions $\{\textsc{Left}, \textsc{Forward}, \textsc{Right}, \textsc{Reverse} \}$, where forward corresponds to commanding $\SI[per-mode=symbol]{3.0}{\meter\per\second}$ along the vehicle longitudinal axis, left and right correspond to commanding $\SI[per-mode=symbol]{3.0}{\meter\per\second}$ with a $\SI[per-mode=symbol]{15}{\degree}$ yaw to the left or right of the vehicle longitudinal axis, and reverse corresponds to commanding $\SI[per-mode=symbol]{-0.5}{\meter\per\second}$ along the longitudinal direction. 
The UAV was restricted to 2D-flight at a fixed altitude above ground. 
These constraints made solving the obstacle avoidance problem much more difficult and provides a better test case for the learned system.
While this small action set may seem limited, the $\SI[per-mode=symbol]{10}{\Hz}$ update rate and the use of a low-level controller enable a rich variety of maneuvers to be conducted.

Early in this work, we tested a stochastic policy neural network that directly output actions from a continuous domain.
We trained this network with Proximal Policy Optimization~\cite{schulman2017proximal} and found that training was much more difficult and resulted in trajectories with frequent small changes in commanded velocity.
The performance of the selected Q-network architecture was higher and the resulting trajectories were observed to be much smoother. 

The network was trained to reach a goal location as quickly as possible without collision.
The reward function was defined as 
\begin{equation} 
    r_t(s_t, a_t) =
    \begin{cases} 
      x_t - x_{t-1} & \textsc{NoCollision}(s_t) \\
      -\lambda & \textsc{Collision}(s_t) 
   \end{cases}
\end{equation}
where $s_t$ is the complete vehicle state, $a_t$ is the action commanded, $x_t$ is the vehicle's x-position at time-step $t$ (with the x-axis as the road direction), 
and $\lambda$ is a penalty for collision or out-of-bounds states. 
If the vehicle has not crashed or flown out-of-bounds, the reward is equal to the forward progress made since the previous time-step. 
If the vehicle has crashed or flown out of bounds, the reward is equal to the penalty $-\lambda$. 

\begin{figure}[htb]
    \centering 
    \includegraphics[width= 0.45\textwidth]{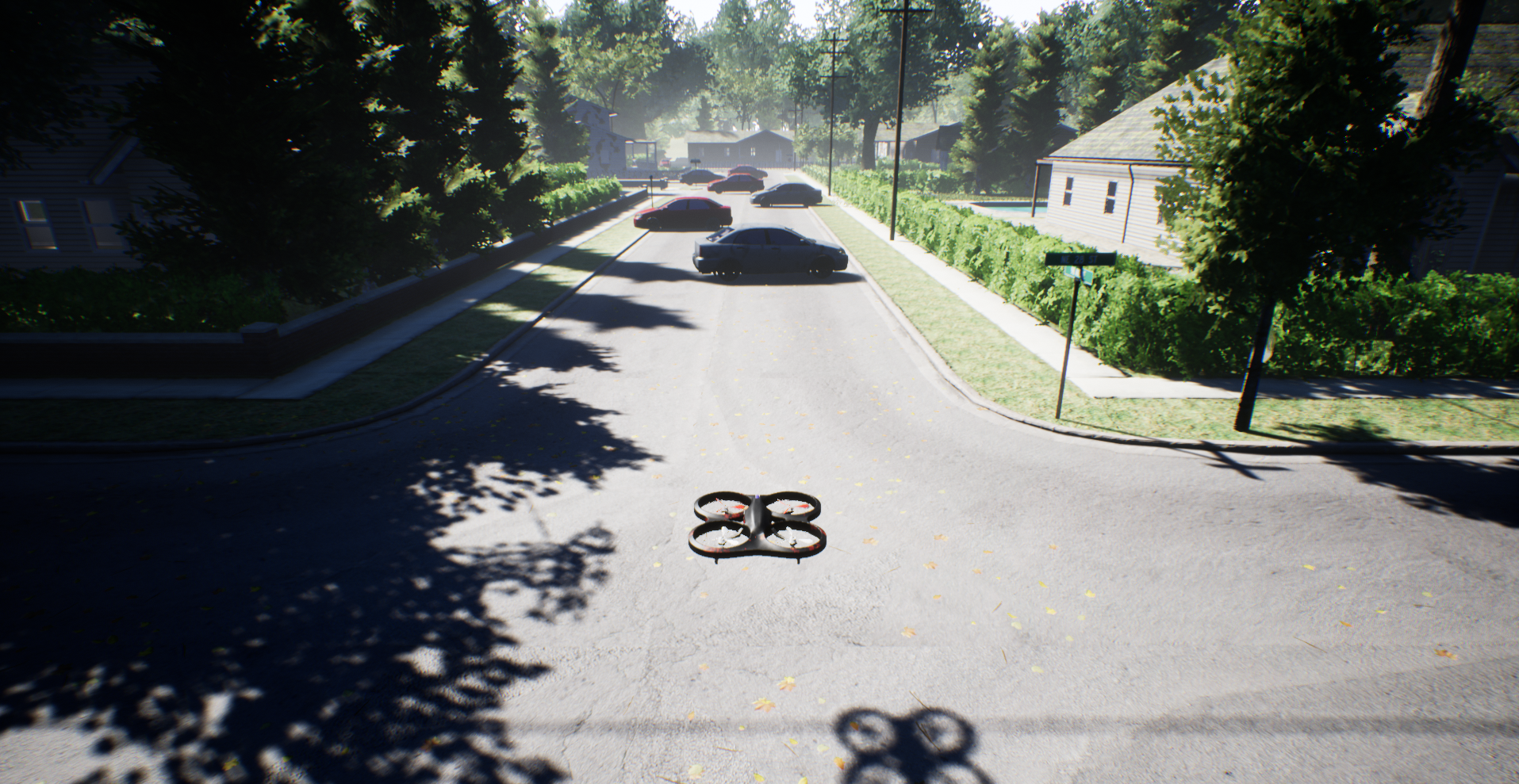} 
    \adjincludegraphics[width= 0.45\textwidth, trim={0 0 0 {.105\height}},clip]{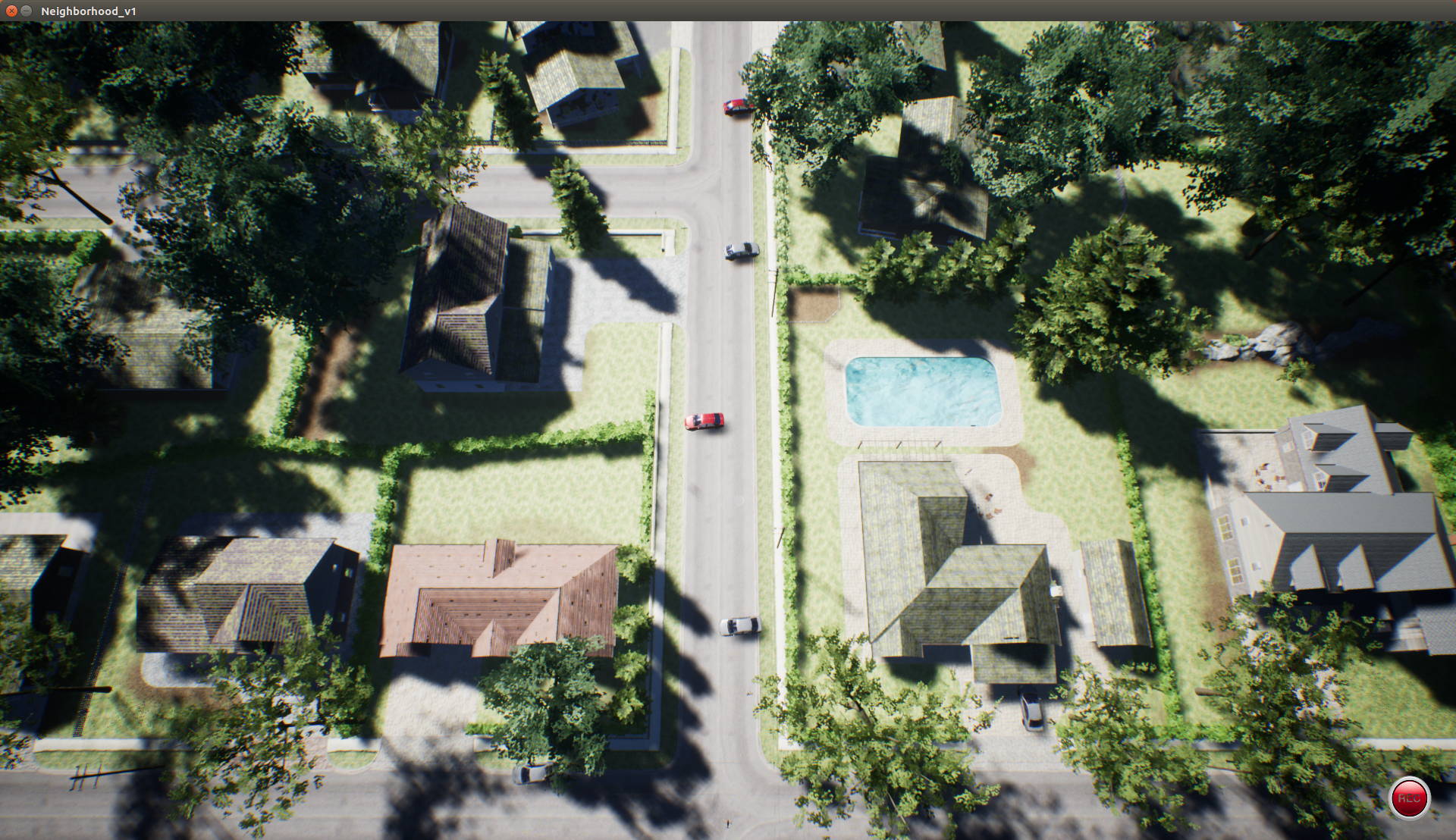}
    \caption{UAV obstacle course. The left image shows the obstacle course from just behind the UAV point-of-view. The right shows a top-down view of the course. The UAV must fly forward while avoiding the cars and staying above the roadway to successfully navigate the course. \label{fig:obstacles}}
\end{figure}

The policy was trained to navigate an obstacle course we created in the AirSim Neighborhood environment. 
The UAV's goal was to fly from one end of a street to the other.
Cars were placed as obstacles in the street and the UAV was constrained to fly at an altitude approximately equal to half the height of the cars.
Images of this obstacle course are shown in~\cref{fig:obstacles}.

The course is 100 meters long with six cars placed as obstacles. 
The cars were spaced evenly from the 20 meter mark to the 100 meter mark at 15 meter intervals. 
At the beginning of each episode, the lateral position of each car was randomly assigned. 
The quadrotor was positioned at the 0 meter mark at the beginning of each episode. 
When the quadrotor safely reached the end of the $\SI[per-mode=symbol]{100}{\meter}$ long street, crashed into an obstacle, flew out of bounds, or exceeded 600 time-steps (1 minute) of flight time, the episode was terminated.

The policy was trained using estimates from the depth map network and simulated odometry measurements from AirSim as input. 
The velocity of the maximum Q-value action was commanded at each time-step.
The network was optimized using the RAINBOW algorithm. 

\subsection{Collision Prediction Network}
The input to the first layer of the collision prediction network is the concatenation of the policy network input. 
This input is processed through four fully-connected layers and the resulting hidden vector is concatenated with the selected action index. 
This intermediate layer is further processed through four additional layers, returning a $2 \times 1$ output. 
This output vector represents the un-normalized probabilities of collision versus non-collision in the next $d$ time-steps. 
The normalized probabilities are calculated using the softmax logistic transform.
The architecture is summarized in~\cref{tab:collision_net}.
\begin{table}[ht]
    \centering
    \caption{Collision Network Architecture.}
    \begin{tabular}{@{}cccc@{}}
         Layer & Input Length & Activation & Output Units \\
         \toprule
         Input & 10 & Leaky ReLU & 256 \\
         \midrule
         H1 & 256 & Leaky ReLU & 256 \\
         H2 & 256 & Leaky ReLU & 128 \\
         H3 & 128 & Leaky ReLU & 32 \\
         \midrule
         Concatenate & 32 + 4 & None & 36 \\
         \midrule
         H4 & 36 & Leaky ReLU & 32 \\
         H5 & 32 & Leaky ReLU & 32 \\
         H6 & 32 & Leaky ReLU & 32 \\
         \midrule
         Output & 32 & None & 2 \\
         \bottomrule
    \end{tabular}
    \label{tab:collision_net}
\end{table}

The collision network was trained by supervised learning.
The training data was gathered by running the depth estimation network and policy network in the obstacle course for 1,000 episodes without the collision prediction network.
The depth map estimates and odometry vectors were collected as the training inputs at each time step. 
The AirSim collision flag was collected each time step as well. 
The training labels were set to $(1,0)$ for each time step in which a collision was observed in any of the subsequent ten frames, and $(0,0)$ otherwise.

The training process minimized the cross entropy loss for a binary distribution between the target distribution and the predicted distribution, as in
\begin{equation}
    L = - (y_t \log (\hat{y}_t) + (1 - y_t) \log (1 - \hat{y}_t))
\end{equation}
where $y_t$ is the true probability of collision and $\hat{y}_t$ is the prediction.

The training data set had far more examples of non-collision frames than collision frames, with 295,183 total frames and just 1,890 with collisions occurring in the next ten frames. 
Sampling from this set directly would often result in training batches that incorrectly estimate an expected crash rate of zero, which could bias the training process. 
To offset this, we constructed each training batch by sampling $25\%$ of the batch to be collision examples and $75\%$ of the batch to be non-collision examples.
Sampling in this way could also bias the training to be overly conservative. 
To offset this potential bias, we considered using an importance sampling weighting on each loss component.
Initial testing, however, showed this bias not to be an issue and importance sampling was not used for our results.

\section{Experiments and Results}
The depth network, policy network, and collision prediction network were each trained separately. 
The performance of each network was also evaluated independently using appropriate metrics. 
The best performing instance of each of the three networks was then integrated into the hybrid obstacle avoidance system and the performance of the complete system evaluated.
Systems that only used either the RL policy or the contingency policy were also tested as baselines for comparison. 

\subsection{Depth Network}

A data set of $14,048$ \textit{(RGB image, reduced depth map)} pairs was collected, of which ninety percent were used to train the depth network and ten percent of were held out for performance validation.
The performance of the trained network on the validation set was measured using several different loss functions. 

The mean average error (MAE) is defined as
\begin{equation}
    \text{MAE} = \sum_{i=1}^n\|y_i - \hat{y}_i\|_1
\end{equation}
and evaluates the general fit of the estimates to the data. 
The mean square error (MSE) is defined as
\begin{equation}
    \text{MSE} = \sum_{i=1}^n\|y_i - \hat{y}_i\|^2_2
\end{equation}
and similarly evaluates the data agreement, while putting relatively more emphasis on larger magnitude errors than MAE. 
The root mean-square logistic error (RMSLE) is defined as 
\begin{equation}
    \text{RMSLE} = \sum_{i=1}^n\|\text{log}(2 - y_i) - \text{log}(2 - \hat{y}_i)\|^2_2
\end{equation}
and evaluates the accuracy of estimates relative to the regional intensity of image pixels. 
This means that errors are penalized more heavily in regions with high pixel values than errors of equal magnitude in regions with low pixel values. 

The mean validation loss of the trained depth estimation network for these metrics and the Huber loss are shown along with standard error bounds in~\cref{tab:depth_loss}.
\begin{table}[htb]
    \centering
    \caption{Depth estimation network performance. Disagreement loss calculated by four metrics: mean average error (MAE), mean square error (MSE), root mean-square logistic error (RMSLE) , and Huber loss.}
    \begin{tabular}{@{}cccc@{}}
        MAE & MSE & RMSLE & Huber \\
        \midrule
         $0.147 \pm 1.41 \times 10^{-3}$ & $0.043 \pm 9.01 \times 10^{-4}$ & $0.102 \pm 6.60 \times 10^{-4}$ & $0.022 \pm 4.49 \times 10^{-4}$ \\
         \bottomrule
    \end{tabular}
    \label{tab:depth_loss}
\end{table}

The observed validation error is reasonable based on the performance of similar depth-estimation CNNs reported in prior works~\cite{mern2019}.
We tested the effectiveness of the depth network as part of the complete system by comparing the performance with the depth network estimates to the performance of the system using the true depth map. 
The result of this testing is discussed in the following sections. 

\subsection{Policy Network}
The policy network was initially trained with a collision penalty $\lambda = 20.0$ for 5,199 episodes totalling 983,751 time-steps.
Its performance was evaluated by running the network through the obstacle course for 1,000 trials. 
Since the cars were positioned randomly at the beginning of each episode, it was not necessary to change the obstacle course for evaluation. 
During evaluation, the noise factor in the final layers of the policy network used for exploration was set to zero.
The policy was evaluated on the course completion rate, the collision rate, and average episode length. 

Evaluating the initial network showed a higher collision rate than acceptable. 
We first tried to address this by tuning the collision loss parameter. 
The crash penalty was increased to 30 and the policy network was trained for a further 1,720 episodes. 
This additional training led to a slight increase in the fraction of episodes completed during evaluation. 
However, the collision rate remained unacceptably high.
The crash penalty was also increased to 40 and the policy network was trained for a further 1,234 episodes (336,202 total time-steps). 
This further training decreased the fraction of episodes completed during evaluation.
These results are summarized in~\cref{tab:rl_eval_results}.

In order to evaluate the impact of depth estimation performance on the policy network performance, we also trained and tested a policy network using the true depth maps provide directly by AirSim. 
This network was trained for 4,397 episodes with a collision penalty of $\lambda = 20$ and evaluated for 1,000 trials. 
These results are also included in~\cref{tab:rl_eval_results}.
The comparable completion rate when using the true depth network suggests that the depth estimation network performance is sufficient for the collision avoidance task.
The increased out-of-bounds rate with the true depth maps suggests the true maps may have caused difficulties with training the policy, resulting in convergence to a lower-performing local optimum.

\begin{table}[htb]
    \centering
    \caption{Policy network evaluation results. Results from 1,000 test runs of the policy network without using the collision detection network. The collision, out-of-bounds (OOB), and timeout rates are shown. The mean number of time-steps per episode and one standard error bounds are also shown. Each time-step is 0.1 seconds in simulation time.}
    \begin{tabular}{@{}cccccccc@{}}
         $\lambda$ & Depth Map & Collision Rate & OOB Rate & Timeout Rate & Completion Rate & Episode Length \\
        \toprule 
          20.0 & Estimated & $17.7 \%$ & $2.4 \%$ & $0 \%$ & $79.9 \%$ & $286.5 \pm 3.6$ \\
          30.0 & Estimated & $18 \%$ & $0.09 \%$ & $0 \%$ & $81.1 \%$ & $295.8 \pm 2.7$ \\
          40.0 & Estimated & $18.5 \%$ & $ 17.7\%$ & $ 4.0\%$ & $59.7 \%$ & $321.4 \pm 3.6$ \\ 
          \midrule
          20.0 & True & $9.4 \%$ & $24.5 \%$ & $0.0 \%$ & $66.0 \%$ & $308.2 \pm 2.4$ \\
          \bottomrule
    \end{tabular}
    \label{tab:rl_eval_results}
\end{table}
\subsection{Collision Prediction Network}
Results from the policy network evaluation showed that tuning the collision loss parameter was not sufficient to achieve a low collision rate. 
To evaluate the effectiveness of the collision prediction network, we assessed its error rates on trajectories generated from the UAV flown by the depth estimation and policy networks. 
In this way, the test set was generated in the same way as the training set.
The data set contained 189 collision-labeled frames and 29,000 collision-free frames. 
A total of 25,000 batches were analyzed.

Because there were disproportionately more non-collision frames, the evaluation was run in batches where each batch was sampled to contain 24 non-collision frames and 8 collision-labeled frames. 
From the results, we calculated the accuracy, precision, recall, and $\text{F}_1$ score. 
The results are summarized in~\cref{tab:collision_net}.
The false positive and false negative errors are both on the order of $1\%$. 

\begin{table}[ht]
    \centering
    \caption{Collision Prediction Network Validation Results.}
    \begin{tabular}{@{}lr@{}}
         Metric & Value \\
         \toprule
         Accuracy & 0.982 \\
         Precision & 0.976 \\
         Recall & 0.954 \\
         F1-score & 0.963 \\
         Cross Entropy & 0.057\\
         True Positive Rate & 0.239 \\
         True Negative Rate & 0.743 \\
         False Positive Rate & 0.006 \\
         False Negative Rate & 0.012 \\
         \bottomrule
    \end{tabular}
    \label{tab:collision_pred_results}
\end{table}

\subsection{Hybrid System}\label{sec:system_testing}
The best performing instance of each of the three networks was integrated into the complete obstacle avoidance system. 
In the case of the policy network, this is the network that was trained with the collision penalty $\lambda = 40$. 
If the collision prediction network predicted that a collision would occur within the next timesteps with greater than 50\% probability, then the contingency policy would take control of the aircraft. 
The contingency policy would then pilot the aircraft until it terminated, at which point control of the aircraft would be returned to the RL controller. 
Systems using both $A^*$ and the expert policy were each tested. 

As a baseline, we compare the performance of our hybrid system with the performance of the best performing policy network alone (without activating the contingency policy). 
These results are shown in~\ref{tab:rl_eval_results} and are reproduced in~\ref{tab:complete_system_results} for convenience. 
We also compare our hybrid system against a system that only uses the contingency policy without the learned RL component. 
In this case, the aircraft simply flies forward in a straight line until the collision detection network goes off. 
At this point, the expert contingency policy takes control of the aircraft. 
After the contingency policy finishes executing, the aircraft resumes flying forward in a straight line. 

Because the dynamics of the aircraft differ when the it is being flown forward in a straight line than when it is piloted by the RL policy, the collision network trained with RL policy data would perform more poorly during straight line flight. 
Because of this, we trained a separate collision prediction network for use when the aircraft is flown forward in a straight line. 
This network was trained in the same manner to the original collision prediction network using data gathered from straight line flight. 
This network achieved similar validation results to the RL policy prediction network. 

The performance of each system was evaluated on the obstacle course for 1,000 episodes. 

\begin{table}[ht]
    \centering
    \begin{threeparttable}
    \caption{Hybrid System Evaluation Results.}
    \begin{tabular}{@{}ccccc@{}}
          & Hybrid ($A*$) & Hybrid (expert) & RL policy & Expert Only \\
         \toprule
         Collision Rate & $5.5\%$ & $1.8\%$ & $17.2\%$ &  $24.5\%$  \\
         Contingency Policy Collision Rate & $10.1\%$ & $3.3\%$ & -- & $36.1\%$ \\
         Out-of-Bounds Rate & $0.1\%$ & $0\%$ & $0.9\%$ & $0\%$ \\
         Timeout Rate & $0.1\%$ & $0\%$ & $0\%$ & $0.2\%$ \\
         Completion Rate & $84.2\%$ & $94.9\%$ & $81.9\%$ & $39.2\%$ \\
         Episode Length* & $468.2 \pm 5.4$ & $440.2 \pm$ $3.2$ & $332.8 \pm 0.1$ & $617.2 \pm 7.8$ \\
         \bottomrule
    \end{tabular}
    \label{tab:complete_system_results}
    \begin{tablenotes}
      \small
      \item *Episode lengths are calculated only for episodes that end in completion.
    \end{tablenotes}
    \end{threeparttable}
\end{table}

The hybrid system using the expert contingency policy achieved a completion rate of $94.9\%$. This is significantly higher than the completion rates achieved by the RL policy alone or the expert contingency policy alone. 
The hybrid system using $A^*$ also achieved a higher completion rate than those of RL policy alone and the expert contingency policy alone. However, its completion rate was significantly lower than that of the hybrid system using the expert policy. 

The hybrid system using the expert contingency policy achieved an average completed episode length of $440.2$ timesteps. 
Although this is longer than average completed episode length of $332.8$ achieved by the RL policy alone, it is significantly shorter than the average episode length of $617.2$ achieved by the expert contingency policy alone. 
This shows that our hybrid approach is able to balance the objective of quickly reaching its destination with the objective of safely avoiding collisions. 

Qualitative analysis of the hybrid system using $A^*$ shows that many of its collisions occurred when the $A^*$ planner planned a path that ran too close to the obstacle it was trying to avoid. 
This is likely due to the fact that the $A^*$ planner relies on having an accurate map of its surroundings in order to plan a safe route.
In our system, these maps are generated using the depth estimation network.
Noise in these estimates can have significant negative impacts on the paths planned by the $A^*$ planner.

In contrast, the expert contingency policy relies very little on the depth estimates generated by the depth estimation network.
This is likely a major reason why the hybrid system using the expert policy achieved much a higher completion rate than that achieved by the hybrid system using $A^*$. 
Although the expert policy we designed was very effective in this environment, it is a less general approach than using the $A^*$ planner. 

Examination of the hybrid system using the expert contingency policy shows that, of all the collisions that occurred while the RL policy was piloting the aircraft, the vast majority took place immediately after control of the aircraft was returned to the RL policy from the contingency policy. Almost no collisions occurred as a result of the collision prediction network failing to activate the contingency policy. 
This indicates that the majority of the collisions from our hybrid system are due to weaknesses in the contingency policy. In many instances of collision, the contingency policy would pilot the aircraft nearly around an obstacle but not quite clear it. The RL policy would then hit the corner of the obstacle that the contingency policy failed to completely clear. Presumably, if this expert contingency policy were replaced with a more carefully designed expert policy, then these collisions would be avoided. However, the objective of this paper is not to design an optimal contingency policy for use on this environment. Instead, we show that our hybrid system, when used with an appropriate contingency policy, achieves much better performance than do either the RL policy or the contingency policy alone. 


\section{Conclusions}

This work presented an obstacle avoidance system for small UAVs based on a hybrid neural network and path planner controller using monocular image inputs. 
The proposed hybrid system combines the strength of a learned RL policy and a path planning or expert contingency policy. 
By decomposing collision avoidance in this way, our hybrid system achieves significantly lower collision rates than either the RL policy or the contingency policy do alone. 
Our hybrid system also reaches its destination faster than the contingency policy alone does, demonstrating that it is able to quickly reach its destination while also safely avoiding collisions. 
Our system balances these dual objectives without relying on extensive tuning of the RL reward function.  

The complete system is composed of three specialized neural networks for depth estimation, piloting, and collision detection. 
Experiments showed that neural network piloting performed as well with estimated depth as with the true depth maps. 
This suggests that the proposed depth network was sufficiently accurate for the guidance task. 
However, the $A^*$ planner used as a contingency policy seemed to perform poorly due to the estimate noise. 
In practical applications, even small UAVs may have short range obstacle sensors that would be able to provide better map data for contingency systems.
Future work will model this data and investigate its effect on system performance.

The reinforcement learning trained neural network controller was able to reach the goal more quickly than the expert policy on completed episodes. 
Qualitative inspection suggests this is due to the RL policy following much smoother trajectories than the expert. 
However, the collision rate of the RL policy alone was unacceptably high. 
Increasing the penalty for collisions during RL training was not sufficient to resolve this. 

The collision detection network performed well, with type I and type II errors each on the order of $1\%$. 
These errors may be higher when used on control polices that differ for the one on which it was trained. 
Future work will investigate this and robust training solutions using curricular replay buffers.

The focus of this work was to demonstrate that a hybrid approach to collision avoidance for small UAVs can outperform approaches that rely on RL, expert policies, or path planning algorithms alone. 
Replacing the contingency policies that we used in our experiments with more sophisticated path planning algorithms will likely further reduce the collision rate. 
Future work will explore more robust contingency policies, for example using uncertainty-aware Monte Carlo tree search or Hamilton-Jacobi reachability. 
\bibliography{bibliography}

\appendix
\section*{Appendix}
\subsection{Expert Contingency Policy}
The expert contingency policy we used in these experiments operates as follows: 

\begin{algorithm}[H]
\SetAlgoLined
Take control of the aircraft from the RL policy.

Slow quadrotor to a stop.

Take an RGB image with the on-board camera and generate a depth map using the depth estimation network.

\eIf{the aircraft is to the right of the middle of the track} {
    \eIf{the left half of the middle three rows of the depth map do not contain any cells closer than 10 meters} {
        Fly the aircraft right until it reaches within 0.5 meters of the right boundary of the track.
    } {
        Fly the aircraft left until it reaches within 0.5 meters of the left boundary of the track.
    }
} {
    \eIf{the right half of the middle three rows of the depth map do not contain any cells closer than 10 meters} {
        Fly the aircraft left until it reaches within 0.5 meters of the left boundary of the track.
    } {
        Fly the aircraft right until it reaches within 0.5 meters of the right boundary of the track.
    }
}

Return control of the aircraft to the RL policy.

\caption{Expert Contingency Policy}
\end{algorithm}

\subsection{$A^*$ Planner Contingency Policy}
The $A^*$ contingency policy used the $A^*$ path planning algorithm to route the UAV around detected obstacles. 
A rules-based policy was used to orient the vehicle before executing $A^*$ and to handle the input and output from the search.
The policy procedures are described here. 




    

First, 
the contingency policy slows the quadrotor to a stop. 
\textit{Procedure A} is then executed. 
If \textit{Procedure A} returns that there are no obstacles detected, then the aircraft is flown forward for 4 meters and control is then returned the RL policy. 
If \textit{Procedure A} returns a path, it will pilot the aircraft along this path, and then return control of the aircraft to the RL policy. 

If \textit{Procedure A} finds that there are no paths around the obstacle, then the aircraft is rotated right by 30 degrees. 
\textit{Procedure A} is then called again. 
If \textit{Procedure A} returns that there are no obstacles detected, then the aircraft is flown forward in the current direction its facing for 4 meters or until the aircraft reaches within 0.5 meters of the boundary of the track. 
Control of the aircraft is then returned to the RL policy. 
If \textit{Procedure A }returns a path, it will pilot the aircraft along this path, and then return control of the aircraft to the RL policy. 

If \textit{Procedure A} again finds that there are no paths around the obstacle, then the aircraft is rotated back to its original orientation, and then is rotated left by 30 degrees and the above process is repeated. 

If \textit{Procedure A} finds that there are no paths around the obstacle, then the aircraft is rotated back to its original orientation, and backed up for 1 meter. 
The steps described in the previous three paragraphs are then repeated. 
If still no path is found at this point, then control of the aircraft is returned to the RL policy. 

\subsubsection{Planning Procedures}
\textbf{Procedure A:} First, an RGB image is taken with the on-board camera and a depth map is generated using the depth estimation network. A $(1 \times 16)$ occupancy map is then created using the depth map via the process described in \textit{Procedure B}.   If no cells in the occupancy map are marked as occupied, the procedure returns that no obstacles are detected. Otherwise, it will generate an obstacle map using the process described in \textit{Procedure C}. Once an obstacle map has been generated, it will run $A^*$ to try to find a route past the obstacle using the process described in \textit{Procedure D}. If a path is found, this path is returned. If no path is found, then "no path found" is returned. 

\textbf{Procedure B:} The occupancy map is created by looking at the middle three rows of the depth estimate (recall that the depth estimates have the shape $(9 \times 16)$. If any column in the three middle rows of the depth estimate contains a cell closer than 10 meters, then the corresponding column in the occupancy map is marked as occupied. Otherwise, it is marked as vacant.

\textbf{Procedure C:} 
First, the lines of sight from the on-board camera are projected out as lines going forward at 45 degrees from either side of the aircraft.
For each contiguous block of occupied cells in the occupancy map, the minimum distanced contained in the corresponding columns of the three middle rows of the depth map is found. The width of each of these cells is found by calculating how wide the camera's field of vision is at the distance found previously, and then dividing this width by 16 (the number of cells in the occupancy map). An obstacle is then represented as a square, with its nearest face being located at the minimum distance found earlier. The width of the square is equal to total the width of the contiguous block of occupied cells in the occupancy map at the aforementioned distance. The square extends back to a depth equal to its width. 

This process of generating obstacles is repeated for each contiguous block of occupied cells in the occupancy map. 
The positions and orientations of the squares are adjusted according to the angle at which the aircraft was facing when the obstacle map was generated. 

\textbf{Procedure D:} 
First, a rectangle is defined, the width of which is equal to the width of the flight track and the depth of which is equal to either 5 meters or distance of the furthest corner of any obstacle on the obstacle map, whichever is greater. 
Then, the lines of sight from the on-board camera are projected out going forward at 45 degrees from either side of the aircraft at the angle at which the aircraft is facing.

The goal location that $A^*$ is trying to reach is defined as any point with a distance equal to or greater than the furthest end of the rectangle. In other words, any point that $A^*$ finds that is past the furthest end of the rectangle counts as a goal location. Additionally, $A^*$ is constrained plan paths which do not go past the left or right lines of the field of vision, and which do not go outside of any of the sides of the rectangle (except for the goal side, which it may pass). 

The $A^*$ algorithm is then used to find a path within the allowable flight space, avoiding any obstacles, to the goal edge. The $A^*$ planner can use any of the following steps for the aircraft when planning a path: fly forward 1 meter, fly 45 degrees forward and to the right 1 meter, fly 90 degrees to the right 1 meter, fly 45 degrees back and to the right 1 meter, fly back one meter, fly 45 degrees back and to the left 1 meter, fly 90 degrees to the left 1 meter, fly 45 degrees forward and to the left one meter.











\end{document}